\documentclass[10pt,twocolumn,letterpaper]{article}

\usepackage{iccv}
\usepackage{times}
\usepackage{epsfig}
\usepackage{graphicx}
\usepackage{amsmath}
\usepackage{amssymb}

\usepackage{booktabs}

\usepackage[pagebackref=true,breaklinks=true,letterpaper=true,colorlinks,bookmarks=false]{hyperref}

\iccvfinalcopy 

\usepackage{multirow}
\usepackage{multicol}
\usepackage{caption}
\usepackage{graphicx}
\usepackage{booktabs}
\usepackage{siunitx} 
\usepackage{color, colortbl} 
\usepackage{pifont}
\usepackage{amssymb}
\usepackage{algpseudocode}
\usepackage{algorithm}

\definecolor{Gray}{gray}{0.93}
\usepackage{arydshln} 
\newcommand\mypara[1]{\vspace{1.0mm}\noindent\textbf{#1}}
\newcommand\fl[1]{{\fontfamily{phv}\selectfont\footnotesize#1}}
\newcommand\nl[1]{{\it``#1''}} 

\newcommand{\OurMethod}{LLM-Planner\xspace}

\usepackage[capitalize]{cleveref}
\crefname{section}{Sec.}{Secs.}
\Crefname{section}{Section}{Sections}
\Crefname{table}{Table}{Tables}
\crefname{table}{Tab.}{Tabs.}

\def\iccvPaperID{} 

\begin{document}

\title{\OurMethod: Few-Shot Grounded Planning for Embodied Agents \\with Large Language Models}

\author{Chan Hee Song\\
The Ohio State University\\
{\tt\small song.1855@osu.edu}
\and
Jiaman Wu\\
The Ohio State University\\
{\tt\small wu.5686@osu.edu}
\and
Clayton Washington\\
The Ohio State University\\
{\tt\small washington.534@osu.edu}
\and
Brian M. Sadler\\
DEVCOM ARL\\
{\tt\small brian.m.sadler6.civ@army.mil}
\hspace{-27pt}
\and
Wei-Lun Chao\\
The Ohio State University\\
{\tt\small chao.209@osu.edu}
\and
Yu Su\\
The Ohio State University\\
{\tt\small su.806@osu.edu}
}

\maketitle
\begin{abstract}

This study focuses on using large language models (LLMs) as a planner for embodied agents that can follow natural language instructions to complete complex tasks in a visually-perceived environment.
The high data cost and poor sample efficiency of existing methods hinders the development of versatile agents that are capable of many tasks and can learn new tasks quickly.
In this work, we propose a novel method, \OurMethod, that harnesses the power of large language models to do few-shot planning for embodied agents.
We further propose a simple but effective way to enhance LLMs with physical grounding to generate and update plans that are grounded in the current environment.
Experiments on the ALFRED dataset show that our method can achieve very competitive few-shot performance: Despite using less than 0.5\% of paired training data, \OurMethod achieves competitive performance with recent baselines that are trained using the full training data.
Existing methods can barely complete any task successfully under the same few-shot setting.
Our work opens the door for developing versatile and sample-efficient embodied agents that can quickly learn many tasks. \footnote{Website: \url{https://dki-lab.github.io/LLM-Planner/}}
\end{abstract}
\section{Introduction}

\begin{figure}[ht]
    \centering
    \includegraphics[width=\linewidth]{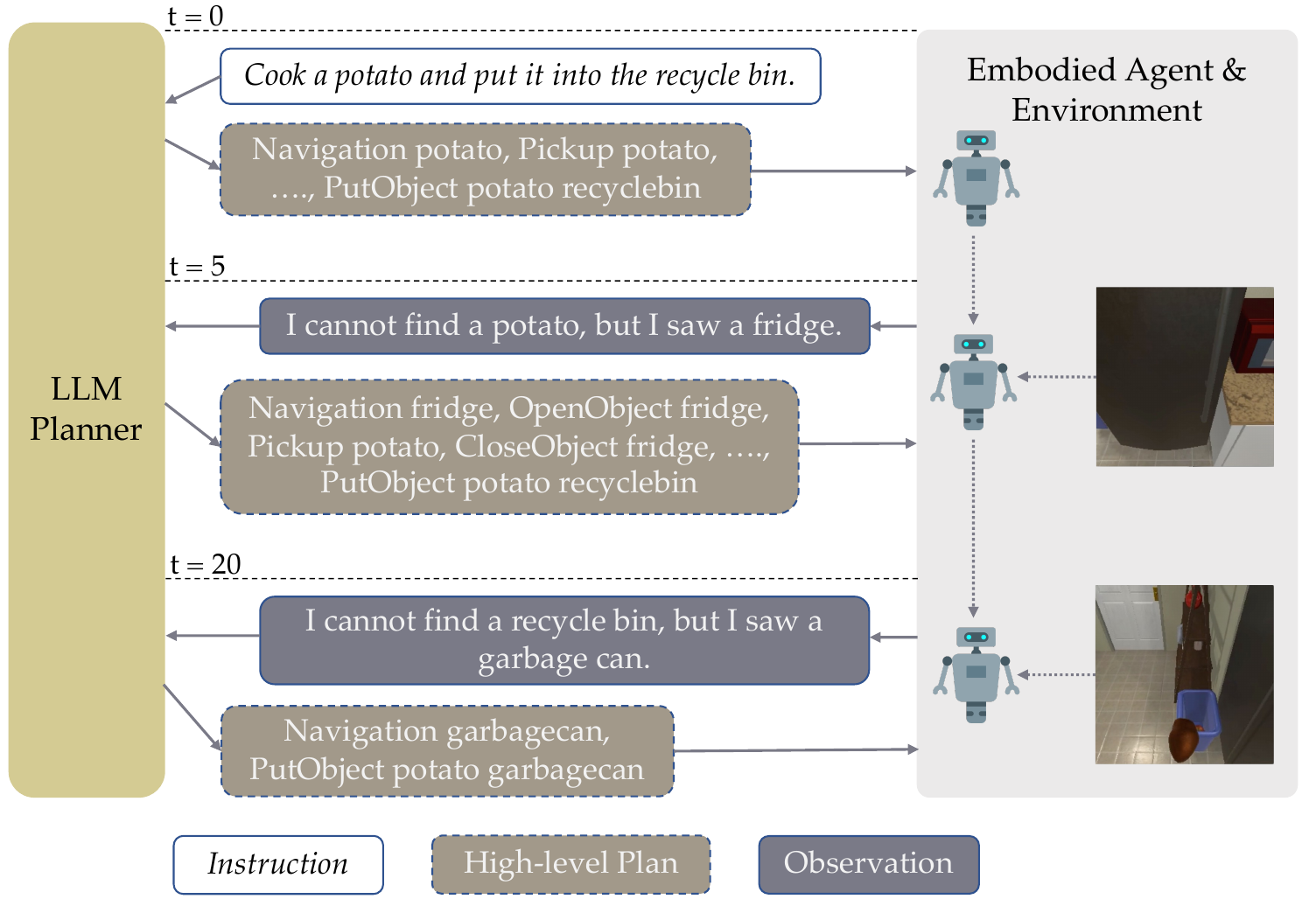}
    \caption{An illustration of \OurMethod for high-level planning. After receiving the natural language instruction ($t=0$), \OurMethod first generates a high-level plan by prompting a large language model (\eg, GPT-3). When the embodied agent gets stuck during the execution of the current plan ($t = 5$ and $20$), LLM-Planner re-plans based on observations from the environment to generate a more grounded plan, which may help the agent get unstuck. The commonsense knowledge in the LLM (\eg, food is often stored in a fridge) allows it to produce plausible high-level plans and re-plan based on new environmental perception.}
    \vskip -5pt
    \label{fig:intro}
\end{figure}

Building versatile embodied agents such as robots that can follow natural language commands to do different tasks as well as learn to do new tasks quickly has long been desired. 
However, contemporary language-driven agents still require a large number of labeled examples (pairs of language instructions and gold trajectories) to learn each task, which is highly costly and hinders the development of truly versatile agents~\cite{ALFRED20, virtualhome, majumdar2020vlnbert, fried2018speaker, https://doi.org/10.48550/arxiv.2108.04927, li-etal-2019-robust, zhang-chai-2021-hierarchical, pashevich2021episodic, Hong_2021_CVPR, r2r, rxr}.
Recently, an array of seminal work has shown the remarkable potential of large language models (LLMs) such as GPT-3~\cite{gpt3} as a few-shot planner for embodied AI agents~\cite{saycan2022arxiv, huang2022prompt, liu2022planning, singh2022progprompt}. 
Agents equipped with LLM-based planners have started to show the ability to learn a new task with a few training examples.

While showing great promises as proof of concepts, existing work still presents significant limitations that may prevent larger-scale applications beyond their limited evaluation setting. 
As an example, SayCan~\cite{saycan2022arxiv}, one of the pioneering work on using LLMs for embodied instruction following, is evaluated on two environments with only 15 object types.
The agent is assumed to be able to enumerate all admissible skills (\ie, [action, object] pairs) up front so it can use an LLM to rank the skills.
This assumption could break easily in partially-observable environments when deploying an agent to new environments.
The cost could also quickly pile up in more complex environments with more objects because the agent needs to call the LLM to evaluate every admissible skill at every step; efficiency deteriorates at the same time. 
Finally, most existing work~\cite{saycan2022arxiv, singh2022progprompt, huang2022prompt, lu_neurosymbolic_planner} uses LLMs to generate a single static plan from the language instruction and then executes on the entire plan.
However, the optimal plan for the same language instruction is dependent on the environment; different environments may need different plans. 
There lacks a way to dynamically adjust the plan from LLMs based on environmental perception.

Building on existing work, we propose \textit{\OurMethod}, an LLM-based planner for embodied instruction following. 
An important design goal is to be able to handle a wide range of tasks in diverse, partially-observable environments, and can dynamically adjust the plan based on perceptions from the environment.
Therefore, different from SayCan, we use LLMs to directly generate plans instead of ranking admissible skills, obviating the need to have sufficient knowledge about the environment \textit{a priori} while also significantly reducing the number of calls to LLMs.
Another unique strength of \textit{\OurMethod} is its ability to dynamically re-plan based on what the agent observes in the current environment, which produces more grounded plans.

More specifically, we adopt hierarchical planning models (e.g., \cite{sutton1999between,sharma-etal-2022-skill}), which consist of a \textit{high-level planner} and a \textit{low-level planner}. 
We use LLMs to generate high-level plans (HLPs), \ie, a sequence of subgoals (\eg, [\fl{Navigation potato}, \fl{Pickup potato}, \fl{Navigation microwave}, ...]) that the agent needs to achieve, in the specified order, to accomplish the final goal specified by the language instruction.
The low-level planner then maps each subgoal into a sequence of primitive actions for achieving that subgoal in the current environment and state.
An important observation is that, \textit{given a high-level plan, low-level planning becomes conditionally independent of the natural language instruction}. It becomes the classic object localization and navigation problem~\cite{robotics_survey} (for navigation subgoals) or simply executing the specified interaction action with the right objects (for interaction subgoals). 
The low-level planner can be trained with data synthesized from the simulator (see, \eg,~\cite{min2022film, blukis2021a}).

Furthermore, we follow the in-context learning paradigm~\cite{gpt3, suvery_prompt} and only use a small number of paired examples.
In addition, no parameter update is needed, which saves development time.
For the example in~\autoref{fig:intro}, at the beginning of an episode ($t=0$), given a natural language instruction, we directly prompt the LLM to generate the HLP by giving it several exemplar pairs of (instruction, HLP) in its context.
We also leverage established techniques such as dynamic in-context example retrieval~\cite{perez2021true,schick2021s,gao-etal-2021-making,liu-etal-2022-makes} and logit biases~\cite{bernal} to further improve the in-context learning performance.

While the HLPs generated by LLMs are already plausible at first glance, they still lack a fundamental aspect of embodied agents --- \textit{physical grounding}; \ie, the generated HLP needs to be grounded to the environment the agent is in. 
Previous approaches~\cite{saycan2022arxiv, singh2022progprompt, huang2022prompt} train a separate model that translates the LLM plans to the grounded admissible actions.
However, this is possible under the assumption that the LLM plan can be matched to a reasonable admissible action.
If the LLM plans are not contained in the list of admissible action, which is the case in the diverse environments, this creates an undetermined behavior for those agents.
To overcome this problem, we propose a novel \textit{grounded re-planning} algorithm to empower \OurMethod with physical grounding.
Specifically, as an agent is executing the initial HLP, whenever it has taken too many steps to reach the current subgoal or has made too many failed attempts, we dynamically prompt the LLM again to generate a new continuation of the partial HLP that has been completed at that point. 
For grounding, we add the list of objects perceived in the environment so far into the prompt as a simple but effective description of the current environment.
\autoref{fig:intro} demonstrates how our grounded re-planning algorithm can help the agent overcome a plan that is unattainable.
For the example at $t=5$, the agent is taking too long to find a potato. 
It re-prompts the LLM with the object \fl{fridge} observed in the environment, and \OurMethod generates a new HLP from scratch (because no subgoal has been completed so far) that directs the agent to look for a potato in the fridge.
By introducing a way to incorporate feedback from the environment, we aim to create a \textit{closed-loop} between the environment and the LLMs where LLMs can dynamically adapt the generated high-level plans to the environment.

While most existing work~\cite{saycan2022arxiv, huang2022inner, huang2022prompt, singh2022progprompt, lu_neurosymbolic_planner} is evaluated under a limited setting (\eg, limited/known environments, short-horizon tasks, or simple environments with a small number of objects), we evaluate \OurMethod on ALFRED~\cite{ALFRED20}, a large-scale dataset with diverse partially-observable environments and a wide variety of tasks and objects. 
We test our \OurMethod by integrating it with the perception module and low-level planner from a strong baseline model, HLSM~\cite{blukis2021a}. 
Using less than \num{0.5}\% of paired training data, \OurMethod achieves competitive performance compared with HLSM and outperforms multiple other baselines, which are trained with the full training set.
Under the same few-shot setting, existing methods can barely complete any task successfully. 
Our work opens a new door for developing versatile and extremely sample-efficient embodied agents by harnessing the power of large language models and grounding.

\section{Related Work}

\subsection{Vision-and-language Navigation} 
In navigation-only VLN datasets such as R2R~\cite{r2r}, models that generate the action sequence end-to-end with a Transformer model can already achieve a good performance~\cite{https://doi.org/10.48550/arxiv.2108.04927, pashevich2021episodic}. 
Recent work~\cite{li-etal-2019-robust, lu2019vilbert,majumdar2020vlnbert,Hong_2021_CVPR} employs BERT and its variants~\cite{DBLP:conf/naacl/DevlinCLT19, liu2020roberta} to get better language understanding. 
These models jointly learn the linguistic and visual representations with cross-attention for grounding. 

However, in more complex VLN, or embodied instruction following in datasets such as ALFRED~\cite{ALFRED20}, hierarchical planning models~\cite{blukis2021a, min2022film, Liu2022LEBPL} that separate the high-level and low-level planning have proven to be most effective. These models use pretrained language models (\eg BERT) to generate high-level plans and construct a semantic map to guide the agent to find the target objects specified in the high-level plan.  

Recent work has shown that hierarchical planning models are advantageous in the low-data regime. (SL)\textsuperscript{3}~\cite{sharma-etal-2022-skill} uses $10\%$ of ALFRED's training data to learn how to generate natural language subtasks  and then match primitive actions to each subtask. We take this modular approach one step further and propose to use large language models (LLMs) under the few-shot setting.
More discussion of (SL)\textsuperscript{3} is in the supplementary materials.

\subsection{Prompting for VLN} 

The use of LLMs for decision making has become an increasingly popular topic for research. 
Two major branches of LLM usages among existing works are 1) using the LLM as an auxiliary helper or 2) using the LLM as a planner. We categorize each work into these categories and outline the difference between those works and ours.

\noindent \textbf{LLM as an Auxiliary Helper} This branch of work uses LLM as an auxiliary helper to generate relevant information to help the main model. LM-Nav~\cite{shah2022robotic} prompts LLMs with raw navigation instructions and \num{3} in-context examples to generate a list of landmarks for a vision-language model to infer a joint probability distribution over landmarks and images.
However, we show that LLM can be used for more than an auxiliary information generator and can be used to perform planning while being grounded to the environment.

\noindent \textbf{LLM as a Planner} This branch of LLM usage focuses on the LLM's ability to generate a plan that is executable in the environment directly or indirectly by using a low-level planner. Several studies have explored the usage of LLM as a planner for embodied agents~\cite{saycan2022arxiv, lu_neurosymbolic_planner, huang2022prompt, Zheng2022JARVISAN, huang2022prompt_map, singh2022progprompt}.
Majority of the works assume the availability of admissible actions in the environment and formulate the approach based on that assumption.
Some are due to the underlying evaluation setup~\cite{huang2022prompt, singh2022progprompt, lu_neurosymbolic_planner}, while others try to train a model to predict a list of admissible actions in the environment~\cite{saycan2022arxiv}.
However, such an assumption leads to various implications on practicality: 1) This admissible action list may be hard or infeasible to obtain, especially in partially-observable environments, and 2) the length of the list grows combinatorially \textit{w.r.t.}\ environment complexity (e.g., \# of objects).
In contrast, LLM-Planner is a \textit{generative model}.
It generates the high-level plan without assuming the knowledge of specifics of the current environment, and dynamically refines the plan based on new observations.
To validate our claim, we implement one of the major works, SayCan~\cite{saycan2022arxiv} to our evaluation dataset (ALFRED) and compare the difference in~\autoref{sec:exp}.

Other work~\cite{Zheng2022JARVISAN} that does not evaluate under that assumption uses LLM as a static generator for high-level plans.
However, we take one step further and propose a \OurMethod without the aforementioned assumptions.
\OurMethod is able to ground the LLM to the current environment by using a pre-trained vision model. 
Next, it can directly predict HLP without relying on a list of admissible actions in the current environment. 
Additionally, \OurMethod can perform the aforementioned capabilities while re-planning during the task execution to dynamically adapt the high-level plans to the current environment.
At last, \OurMethod is evaluated on a diverse set of tasks in the ALFRED environment, testing the real-life applicability of our approach.
With careful prompt design and other techniques for better in-context learning, we show that \OurMethod can generate complete and high-quality high-level plans that are grounded in the current environment with a fraction of labeled data.

\begin{figure*}[th]
    \centering
    \includegraphics[width=.85\linewidth]{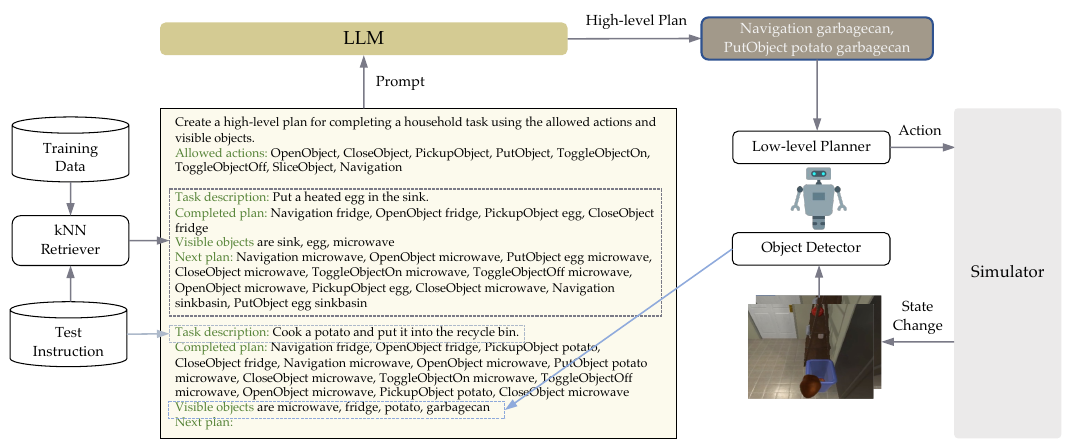}
    \caption{Overview of \OurMethod with prompt design and grounded re-planning.}
    \label{fig:overview}
    \vspace{-10pt}
\end{figure*}

\section{Preliminaries}

\mypara{Vision-and-Language Navigation.} Embodied instruction following is often also referred as vision-and-language navigation (VLN), though it additionally involves interaction actions and usually features a much longer time horizon than typical VLN tasks (\eg, Room2Room~\cite{r2r}). To be consistent with the literature, we will use these two terms interchangeably. 
We will primarily focus on the standard ALFRED~\cite{ALFRED20} dataset, which is built on top of the AI2-Thor~\cite{ai2thor} simulator, but our method can easily generalize to other datasets and environment.
We choose ALFRED mainly considering its diversity in task types (\num{7} different task types) and long-horizon tasks (on average \num{50} actions per task).

The VLN task is defined as following: Given a language instruction $I$, an agent needs to predict and carry out a sequence of primitive actions in the environment $E$ to accomplish the task. In datasets like ALFRED~\cite{ALFRED20}, the instruction $I$ consists of a high-level goal $I_H$ and (optionally) a list of step-by-step instructions $I_L$.
A VLN task can thus be represented by a tuple $(I, E, G)$, where $G$ is the goal test.
We consider hierarchical planning models~\cite{sutton1999between} for VLN, which is explored to various extent in several recent studies~\cite{min2022film, blukis2021a, sharma-etal-2022-skill, Song_2022_CVPR}, but none of them considers the few-shot setting or LLMs for planning.
In this formulation, planning is modeled in a hierarchical fashion. 
The high-level planner maps the instruction $I$ into a high-level plan (HLP) $L_h = \left[g_0, g_1, \cdots, g_T\right]$, where each subgoal $g_i$ is specified as (\fl{high-level action}, \fl{object}).
We define a high-level action to be a collection of primitive actions that can complete a single goal-condition in ALFRED~\cite{ALFRED20}. 
We take the the interaction actions directly from ALFRED and we only add the \fl{Navigation} action.
Therefore, the high-level action space consists of \num{1} navigation action (\fl{Navigation}) and \num{7} interaction actions (\fl{PickupObject, PutObject, OpenObject, CloseObject, ToggleOnObject, ToggleOffObject, SliceObject}).
Similar actions are commonly used in other related work such as SayCan~\cite{saycan2022arxiv} and LM zero-shot planner~\cite{huang2022prompt}.

The low-level planner maps each subgoal  into a sequence of primitive actions $L_l = \left[a_0, a_1, \cdots, a_{T_i}\right]$. State-of-the-art VLN methods~\cite{min2022film, blukis2021a} use a map-based low-level planner and a simple path-finding algorithm to find the target object in the current subgoal from the map.
It is important to note that, once the high-level plan $L_h$ is specified, the low-level planning becomes independent of the instruction $I$.
More formally, $P(L_l | I, L_h, E) = P(L_l | L_h, E)$.
All the components involved in the low-level planner are either deterministic or trained using synthetic data from the simulator.
No paired data involving language instructions is needed.

\mypara{In-Context Learning/Prompting.}
Recently, in-context learning (also known as prompting)\cite{gpt3} has drawn great attention with the rise of LLMs. 
By designing different prompts, LLMs can be adapted to different downstream tasks with a few examples as demonstration without updating any of the parameters. 
In this work, we explore in-context learning with LLMs for embodied agent planning.

\mypara{True Few-Shot Setting.} 
While only using a small number of training examples, many few-shot studies use a large validation set for prompt design and model selection~\cite{gpt3}. 
Recent studies~\cite{perez2021true} have shown that such large validation sets are responsible for overestimation of the efficacy of language models because they create a strong bias for model selection and violate the intended few-shot setting. 
To avoid such bias, we adhere to the true few-shot setting~\cite{perez2021true} in which prompt design and model selection is conducted via cross-validation on the same small training set instead of using a separate validation set.

\section{\OurMethod}

In this section, we describe our method, \OurMethod, which leverages LLMs such as GPT-3 (\textsc{text-davinci-003}) to do few-shot grounded high-level planning for embodied agents.

\subsection{Overview}

LLMs such as GPT-3 are pre-trained to generate natural language.
To adapt them as high-level planners, the first step is to design an appropriate prompt to guide them to generate high-level plans. 
We discuss our prompt design in Section~\ref{sec:prompt}.
The choice of in-context examples is critical for the performance of LLMs, and recent works~\cite{perez2021true,gao-etal-2021-making} have shown that dynamically retrieving similar examples for each test example is beneficial.
We adopt a k-nearest-neighbor (kNN) retriever to select the in-context examples (Section~\ref{sec:retrieval}).
We also use logit biases~\cite{bernal} to further constrain the output space of the LLM to the allowed set of actions and objects.
With all the above designs, we have obtained the \textit{static} version of \OurMethod, which can already generate reasonable HLPs.
In Section~\ref{sec:dynamic}, we propose a novel grounded re-planning algorithm to enhance LLMs with the ability to ground to the current environment, which further improves the HLP quality. 
Finally, we discuss how to integrate \OurMethod into existing embodied agents to empower them with few-shot planning capabilities in Section~\ref{sec:integration}.
An overview of \OurMethod is shown in~\autoref{fig:overview}.

\subsection{Prompt Design}
\label{sec:prompt}

While GPT-3 is shown to be a powerful few-shot learner in a variety of tasks, its power can only be unleashed with carefully designed prompts that are tailored for the desired behavior. 
The final HLP quality can be sensitive to minor design choices in the prompt (\eg, how the HLP is presented, or sometimes even the choice of punctuation).
Therefore, we identify core components of the prompt and systemically compare different design choices under the true few-shot setting based on leave-one-out cross-validation (LOOCV).
The evaluations for some of the key design choices are discussed in Section~\ref{sec:ablation} and~\ref{sec:fine-grained}.

Our final optimal prompt is shown in~\autoref{fig:overview}. 
The prompt begins with an intuitive explanation of the task and the list of allowable high-level actions. 
It is then followed by the in-context examples selected by the kNN retriever (Section~\ref{sec:retrieval}).
When we provide only the high-level goal instruction to GPT-3, we use the format ``Task description: [high-level goal instruction].'' When we include the step-by-step instructions, we include another line ``Step-by-step instructions: [step-by-step instructions]'' following the goal instruction. 
For dynamic grounded re-planning (Section~\ref{sec:dynamic}), we add the subgoals that have been completed and the list of objects observed so far in the environment after the task description. Finally, we append the test example in the same format that ends with ``Next plan:''.

\subsection{In-context Example Retrieval} \label{sec:retrieval}

The in-context examples are an important source of task-specific information for the LLM. 
Different examples could provide different information for the current task.
Intuitively, if the current task is to \nl{cook a potato,} an in-context example that demonstrates the HLP for \nl{cooking an egg} is likely more informative than one that demonstrates how to \nl{clean a plate.}
Specifically, we use a frozen BERT-base model~\cite{DBLP:conf/naacl/DevlinCLT19} to evaluate the pairwise similarity between each training example and the current test example.
The similarity of two examples is defined based on the Euclidean distance between the BERT embedding of their corresponding instruction.
For each test example, we then retrieve the $K$ most similar examples from the small set of paired training examples we have, where $K$ is a hyperparameter that we tune under the true few-shot setting (Section~\ref{sec:fine-grained}).

\begin{algorithm}[t]
\small
\caption{Dynamic Grounded Re-planning with \OurMethod}\label{alg_integration}
\begin{algorithmic}
\State $I \gets \text{Instruction}$
\State $O \gets \text{Set of observed object}$
\State $G \gets \text{List of completed subgoals so far}$

\State $S \gets \text{\OurMethod}(I,O,G)$  \Comment{Full HLP}
\State $t \gets 0$  \Comment{Time step}
\State $k \gets 0$  \Comment{Subgoal index}
\State $s \gets S[k]$ \Comment{First subgoal} 
\State $a_{t} \gets \text{Low-Level-Planner}(s)$ \Comment{First action}

\While{k \textless \text{len}(S)}
    \State execute $a_{t}$
    
    \State $O_t \gets \text{Object-Detector}(\text{current camera input})$
    \State $O\text{.insert}(O_t)$
    
    \If{current subgoal $s$ fails or after $n$ time steps}
    \State $S \gets \text{\OurMethod}(I,O,G)$  \Comment{New HLP}
    \State $k \gets 0$
    \State $s \gets S[k]$

    \ElsIf{current subgoal $s$ is completed}
        \State $k \gets k+1$
        \State $s \gets S[k]$  \Comment{Get next subgoal}
    \EndIf
    
    \State $t \gets t+1$
    \State $a_{t} \gets \text{Low-Level-Planner}(s)$

\EndWhile

\end{algorithmic}
\end{algorithm}
\vskip -10pt

\subsection{Grounded Re-planning}
\label{sec:dynamic}

Using \OurMethod as a \textit{static} high-level planner that only predicts an HLP at the beginning of a task  already shows good data efficiency and accuracy.
As discussed earlier, however, such static planning lacks grounding to the physical environment and can lead to incorrect objects and unattainable plans (\autoref{fig:intro}). 
When such issues happen, the agent cannot complete the current subgoal specified in the HLP, which will lead to one of two possible situations: 1) it fails to execute an action (\eg, bumping into a wall or failing to interact with an object), or 2) it takes a long time and still has not completed the current subgoal (\eg, wandering endlessly).
Intuitively, knowing the objects in the current environment can be very helpful for addressing both of these issues.
For example, knowing that there is a fridge, the LLM may produce an HLP that directs the agent to go to the fridge and try to find a potato in that, because it may have learned the commonsense knowledge that food is likely stored in a fridge during language model pre-training. 

To this end, we present a simple but effective way to enhance LLMs with physical grounding by injecting a list of observed objects, which may be detected using the object detector of the embodied agent, from the environment into the prompt (\autoref{fig:overview}). 
We also add logit biases to these observed objects so \OurMethod can prioritize producing a plan with those objects if they are relevant for the task. 

Based on that, we propose a grounded re-planning algorithm (Algorithm~\ref{alg_integration}) to dynamically update the HLP during the course of completing a task. 
This is in contrast with most existing work that adopts a similar hierarchical planning model (\eg, \cite{min2022film}), which only predicts a fixed HLP up front and sticks to that no matter what happens during the execution.
In our algorithm, re-planning will be triggered under either of two conditions: 1) the agent fails to execute an action, or 2) after a fixed number of time steps.
A new continuation of the already-completed partial HLP will be generated by \OurMethod based on the observed objects, and the agent will carry on with the new plan, which may help it get unstuck.

\subsection{Integration with Existing VLN models}
\label{sec:integration}

We now discuss how to integrate \OurMethod with the existing models to empower them with the few-shot planning capability.
\OurMethod provides a fairly generic and flexible interface for integration. 
As shown in Algorithm~\ref{alg_integration}, it only needs the embodied agent to provide an object list and has a low-level planner that can turn the predicted HLP into low-level actions.
It has no assumption about the inner working of the agent.
For evaluating the end-to-end task completion performance of \OurMethod, we integrate it with a strong baseline method, HLSM~\cite{blukis2021a}, which satisfies such an interface.

\section{Experiments}
\label{sec:exp}

\subsection{Dataset}

We evaluate the efficacy of \OurMethod in generating high-level plans using the ALFRED~\cite{ALFRED20} benchmark, a vision-and-language navigation dataset that requires embodied agents to follow instructions and use visual input to complete tasks in a simulated, spatially continuous household environment. 
The dataset consists of \num{7} task types spanning across 207 unique environments, 115 different object types, and 4,703 tasks.
The task ranges in difficulty from \textit{moving a single object to a new location} to \textit{placing a heated slice of an object into a receptacle}. 
Each task is accompanied by human-written annotations of a high-level goal and a series of more granular step-by-step instructions, created by human annotators as they watched expert demonstrations of the tasks. 
Due to the noise in the natural language instructions and the complexity of planning required to complete such long-horizon tasks, ALFRED is a challenging test of an embodied agent's ability to produce robust and accurate plans. 

\subsection{Metrics} 
We report two main metrics used by ALFRED and one metric created by us to calculate the high-level planning accuracy. 
Success rate (SR) is the percentage of tasks fully completed by the agent. 
A task is only considered complete when all the subgoals are completed. 
Goal-condition success rate (GC) is the percentage of completed goal-conditions. 
Goal-conditions are defined as state changes necessary to complete the task. 
For example, in the task \nl{Slice a heated bread,} bread being sliced and bread being heated are both goal-conditions.

To directly evaluate high-level planning, we introduce a new metric named high-level planning accuracy (HLP ACC), \ie, the accuracy of the predicted HLP compared to the ground-truth HLP. For the static planning setting, we compare the generated HLP with the ground-truth HLP and deems a plan as incorrect if it does not perfectly match the ground truth, and correct otherwise. For the dynamic planning setting, we report a range because we cannot fully separate \OurMethod's performance with the low-level controller choice because we do not have access to an oracle low-level controller. The lower bound is the HLP accuracy of the full generated plan regardless of whether it was executed successfully by the low-level controller (\ie same as evaluating static HLP). The upper bound is the HLP accuracy of the predicted HLP that was successfully executed in the environment by the low-level controller when a task has ended (\ie a task success or a catastrophic failure).

\subsection{Implementation Details}
We choose \num{100} examples for our \OurMethod among 21,023 ALFRED training examples. We apply random stratified sampling to ensure we have a fair representation of all \num{7} task types in the \num{100}-example set. For the kNN retriever, we use the pretrained BERT-base-uncased model from the Huggingface Transformers Library~\cite{wolf-etal-2020-transformers}. For the LLM, we use the public GPT-3~\cite{gpt3} API with \num{9} in-context examples chosen from the \num{100} training examples by the kNN retriever. 
We set the temperature to \num{0} and apply a logit bias of \num{0.1} to all allowable output tokens.
The object list for grounded re-planning is retrieved from the object detector. Specifically, we use the pretrained object detector from HLSM's perception model. We only include objects with a label confidence more than \num{80}\% to reduce noise. It is worth noting that we can potentially use any object detector to obtain the object list, and we only use HLSM's perception model to save computation cost and time. To avoid violating our few-shot assumption, we use the pretrained navigation, perception, and depth model from HLSM which are trained using only synthesized trajectories from the simulator, without any paired training data involving natural language instructions or human annotations.

We compare with two main baseline models, HLSM~\cite{blukis2021a} and FILM~\cite{min2022film}
They are also hierarchical planning models and achieve strong performance on the ALFRED leaderboard.
We directly replace the trained high-level planner for both models with our \OurMethod and did not modify any other parts.
In addition, we re-train these models to compare with \OurMethod under the same few-shot shot setting.
We also compare with several other published baselines models that are trained with the full data. 
Additionally, we also implement SayCan\footnote{\url{https://github.com/google-research/google-research/tree/master/saycan}} to ALFRED and compare under the same few-shot setting as \OurMethod.
Further implementation details can be found in the supplementary.

\label{a:baseline_implementation_details}

\noindent \textbf{SayCan}~\cite{saycan2022arxiv} is a ranking based high-level planner that requires a list of admissible actions and ranks them using the LLM. 
To make it possible for SayCan to work in the complex, partially-observable environments in ALFRED, we give it an \textit{unfair competitive advantage}---it knows all the objects and affordances in the current environment \textit{a priori} to compile the list of skills.
We also equip SayCan with the same kNN retriever from LLM-Planner, which was not needed in their original paper because of the less diverse tasks. More details on the implementation is provided in the supplementary materials.

\noindent \textbf{Other Baselines.} For other baselines included in \autoref{tab_test}, we retrieve the results directly from the published version of the corresponding paper. 
If the ALFRED leaderboard entry is better than the numbers in the original paper, we report the higher.

\begin{table*}[t]
    \centering
    \small
    \tabcolsep 5pt
    \renewcommand\arraystretch{0.9}
    \captionsetup{width=.85\textwidth} 
    \begin{tabular}{lcccccccccccc}
    \toprule
    \multirow{2}{*}{\textbf{Model}} &  \multicolumn{2}{c}{\textbf{Test Unseen}} & \multicolumn{2}{c}{\textbf{Test Seen}} &
    \multicolumn{3}{c}{\textbf{Valid Unseen}} & \multicolumn{3}{c}{\textbf{Valid Seen}}  \\
     \cmidrule(r){2-3} \cmidrule(r){4-5} \cmidrule(r){6-8} \cmidrule(r){9-11}
    &  \textbf{SR}  & \textbf{GC} & \textbf{SR} & \textbf{GC} & \textbf{SR}  & \textbf{GC} & \textbf{HLP ACC} & \textbf{SR} & \textbf{GC} & \textbf{HLP ACC} \\
    \midrule 
    \multicolumn{12}{l}{\textbf{Full-data setting}: \num{21023} (instruction, trajectory) pairs} \\
    \midrule
    \multicolumn{2}{l}{\textbf{Goal instruction only}} \\
    \cmidrule(r){1-1}
    HiTUT~\cite{zhang-chai-2021-hierarchical} & 11.12 & 17.89 &  13.63 & 21.11 & 10.23 & 20.71 & -- & 18.41 & 25.27 & -- \\
    HLSM~\cite{blukis2021a}   & 20.27  & 27.24 & 25.11  & 35.79 & \textbf{18.28} & \textbf{31.24} & 31.24 -- \textbf{70.17} & 29.63 & 38.74 & 38.74 -- \textbf{77.64} \\ \\
    
    \multicolumn{2}{l}{\textbf{Step-by-step instructions}} \\
    \cmidrule(r){1-1}
    E.T.~\cite{pashevich2021episodic} & 8.57 & 18.56 & \textbf{38.42} & \textbf{45.44} & 7.32 & 20.87 & -- & \textbf{46.59} & \textbf{52.92} & -- \\
    HiTUT~\cite{zhang-chai-2021-hierarchical} & 13.87 & 20.31 & 21.27 & 29.97 & 12.44 & 23.71 & -- & 25.24 & 34.85 & -- \\
    M-TRACK~\cite{Song_2022_CVPR} & 16.29 & 22.60 & 24.79 & 33.35 & 17.29 & 28.98 & --  & 26.70 & 33.21 & -- \\
    FILM~\cite{min2022film} & 27.80 & \textbf{38.52} & 28.83 & 39.55 &  -- & -- & 54.93 &-- & --& 60.86 &\\
    LEBP~\cite{Liu2022LEBPL} & \textbf{28.30} & 36.79 & 28.97 & 36.33 &  -- & -- & -- &-- & --&-- & \\
    
    \midrule
    \multicolumn{12}{l}{\textbf{Few-shot setting}: \num{100} (instruction, high-level plan) pairs} \\
    \midrule
    \multicolumn{2}{l}{\textbf{Goal instruction only}} \\
    \cmidrule(r){1-1}

    \OurMethod (Static) + HLSM &  11.58 & 18.47 & 13.05 & 20.58 & 11.10 & 22.44 & 28.67 & 11.82 & 23.54 & 27.45 \\ 
    \OurMethod + HLSM &  13.41 & 22.89 & 15.33 & 24.57 & 12.92 &   25.35  & 33.81 -- 55.85 & 13.53 & 28.28 & 35.08 -- 54.33 \\ \\
    
    \multicolumn{2}{l}{\textbf{Step-by-step instructions}} \\
    \cmidrule(r){1-1}
    HLSM~\cite{blukis2021a} & 0.61  & 3.72 & 0.82  & 6.88 & 0.00 & 1.86 & 0.00  & 0.13 & 2.82 & 0.00 \\
    FILM~\cite{min2022film} & 0.20 & 6.71 & 0.00 & 4.23 & 0.00 & 9.65 & 0.00 & 0.00 & 13.19 & 0.00\\
    SayCan~\cite{saycan2022arxiv} & - & - & - & - & 9.88 & 22.54 & 37.57 & 12.30 & 24.52 & 35.15 \\
    \OurMethod (Static) + HLSM &  15.83 & 20.99 & 17.87 & 23.10 & 14.26 & 26.12 & 43.24 &  15.84 & 25.43 & 39.87  \\ 
    \OurMethod + HLSM & \textbf{16.42} & \textbf{23.37} & \textbf{18.20} & \textbf{26.77} & \textbf{15.36} & \textbf{29.88} & 46.59 -- \textbf{68.31} & \textbf{16.45} & \textbf{30.11} & 50.33 -- \textbf{71.84} \\
     \bottomrule
    \end{tabular}
    \caption{Main results on the ALFRED dataset. "(Static)" means the static planning setting, otherwise it is the default dynamic setting with grounded re-planning. Some methods support using only the goal instruction or additionally using the step-by-step instructions. We compare under both configurations. 
    We could not evaluate SayCan on the test split because ALFRED prohibits using the test metadata, which is needed by SayCan for compiling the admissible actions.}
    \vskip -10pt
    \label{tab_test}
\end{table*}

\begin{table}
  \centering
  \small
  \renewcommand\arraystretch{1}
  \begin{tabular}{lc}
    \toprule
     & \textbf{LOOCV HLP accuracy}  \\
    \midrule
    \textbf{Best Model} &  40.59 \\
    \quad -- \textbf{kNN Retriever} &  17.48 \\
    \quad -- \textbf{Logit Biases} & 38.10 \\
    \quad -- \textbf{Both} & 13.43 \\
    \bottomrule
  \end{tabular}
\caption{Ablation of \OurMethod's components.}
  \label{tab_prompt_ablation}
  \vskip -15pt
\end{table}

\subsection{Main Results}

The main results are shown in~\autoref{tab_test}. We first compare the performance of HLSM when using our \OurMethod as the high-level planner compared with its native version, which is trained using the full training set of ALFRED. 
We find that \OurMethod's few-shot performance is competitive to the original HLSM, and outperforms several recent baselines such as E.T., HiTUT, and M-TRACK, despite using less than \num{0.5}\% of paired training data. 
On the other hand, when trained using the same \num{100} examples (\ie, re-training HLSM's high-level planner), HLSM (and FILM as well) can barely complete any task successfully.
Furthermore, the results show that SayCan still largely underperforms \OurMethod despite the access to the full environment information.
Another significant difference is \textit{cost and efficiency}.
Because of SayCan's ranking nature, it needs to call the LLM many more times than a generative model like \OurMethod: \textit{LLM-Planner calls GPT-3 avg.\ 7 times per task and SayCan calls it 22 times}, even with oracle knowledge of the current environment to shrink the skill list.
Lastly, we see a considerable improvement from grounded re-planning over static planning, especially in the goal instruction only setting, where it improves \num{1.83}\% SR in the unseen test split.
This confirms the effectiveness of the grounded re-planning. 
But we also note that there is still a large room for further improvement.

\begin{figure}[t]
    \centering
    \includegraphics[width=.7\linewidth]{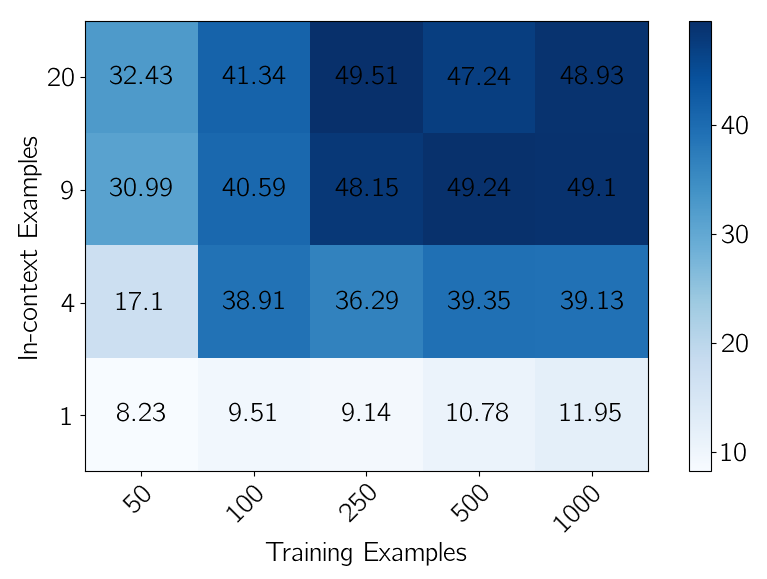}
    \caption{LOOCV HLP accuracy for varying number of in-context examples and training examples.}
    \label{fig:loocv_examples}
    \vskip -10pt
\end{figure}

\begin{figure}[t]
    \centering
    \includegraphics[width=1\linewidth]{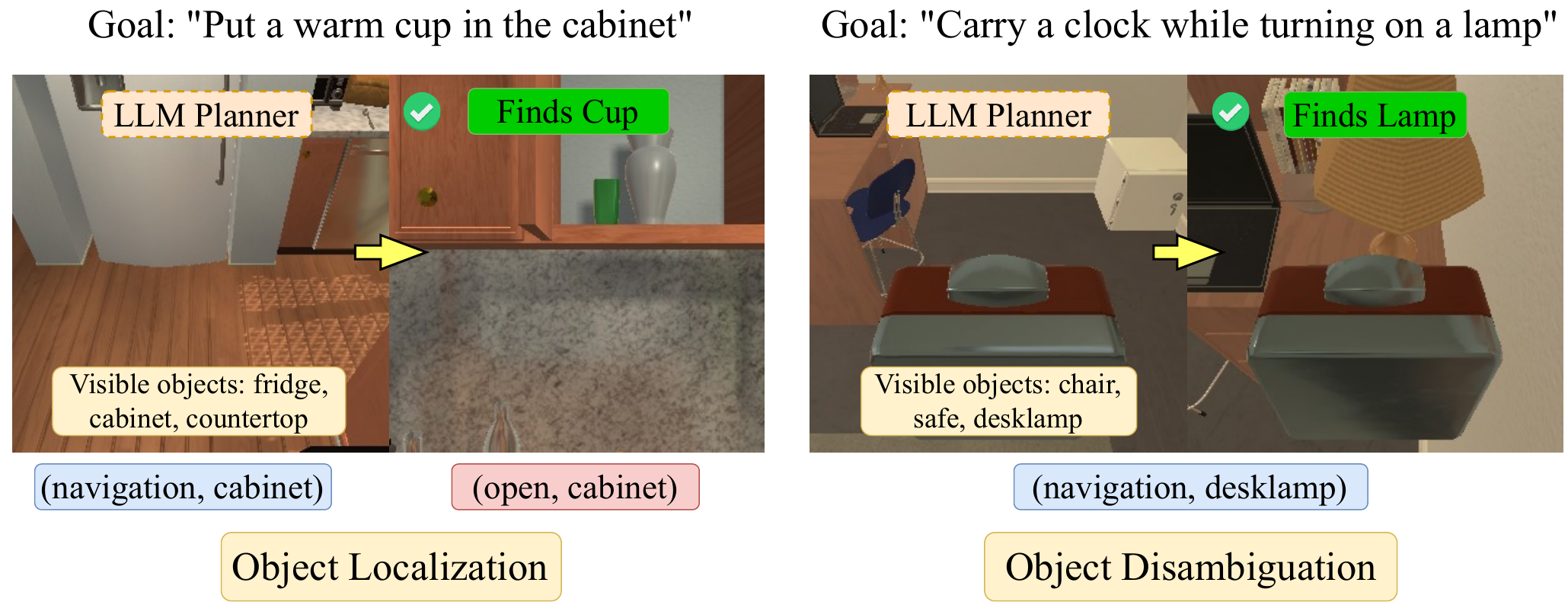}
    \caption{Case studies for \OurMethod.}
    \label{fig:cast_studies}
    \vskip -20pt
\end{figure}

\subsection{Ablation Studies}
\label{sec:ablation}

We conduct an ablation study on different components of \OurMethod to validate their effectiveness. 
We follow the LOOCV process and use only the high-level planning accuracy to determine our choices. Results from this study are in ~\autoref{tab_prompt_ablation}.
We first ablate the kNN retriever module, by replacing it with a retriever that randomly selects in-context examples from the \num{100} example set. Results in~\autoref{tab_prompt_ablation} show that this leads to a significant drop in performance, confirming the necessity of dynamic retrieval. 

Furthermore, we find that enabling logit biases to favor objects that appear in the environment lead to a decent boost in the high-level planning accuracy. Having \OurMethod favor objects that appear in the environment makes it more robust in the cases where the instruction is ambiguous or objects are referred with different names. For example, for an instruction \nl{Turn on the lamp,} different types of lamps, \eg, table lamps or floor lamps, could be. By enabling logit biases to favor objects that appear in the environment (\eg, \fl{TableLamp}), we can correctly guide \OurMethod to output \fl{(TurnOnObject, TableLamp)}. Another example is when the instruction refers to \fl{RecycleBin} but the object name used in the environment is \fl{GarbageCan}. In this case, using logit biases can correctly guide \OurMethod to output the relevant and correct objects.

\subsection{Fine-grained Analyses}
\label{sec:fine-grained}

\mypara{Effect of Number of Examples.} For the main experiments, we chose \num{100} as the number of training examples without any cross-validation because it is our target number for the few-shot setting. 
We then use LOOCV to select the best number of in-context examples using the \num{100} sampled training examples. 
However, we are still curious about the effect of different choices, so we conduct this analysis after the main experiments to show the sensitivity to these hyperparameters. 
It is worth noting that the design choices for the main experiments are not informed by this analysis, to respect the true few-shot setting.

As shown in ~\autoref{fig:loocv_examples}, HLP accuracy generally improves with more training examples, though we start to get a diminishing return around \num{250} training examples. A decent improvement can be expected for the main experiments in Table~\ref{tab_test} if we choose to use more training examples (\eg, \num{250}). Furthermore, we find that \num{9} is generally a good number for in-context examples. Although adding more in-context examples could still improve the performance slightly, it may not be meaningful enough to justify the additional cost. 
Not too surprisingly, more in-context examples is more beneficial when there is less training examples, because there are less useful examples to retrieve from.

\mypara{Case Studies.} In ~\autoref{fig:cast_studies}, we show two examples where \OurMethod helps with object localization and disambiguation through grounded re-planning. For the first case, even using only the high-level goal instruction, \OurMethod correctly predicts that the cup is likely located in the cabinet after failing to find a cup but observing a cabinet in the environment. This shows \OurMethod can achieve a similar effect to what the semantic map tries to achieve in FILM~\cite{min2022film}, \ie, predicting plausible location for target objects. For the second case, we show that \OurMethod can correctly ground the word \nl{lamp} to the \fl{desklamp} in the environment.

\section{Conclusion}

We demonstrate a novel high-level planner based on large language models for embodied agents that can be used in diverse, partially-observable, and complex environments.
It can also dynamically re-plan based on environmental perception to produce more grounded plans.
Our work can dramatically reduce the amount of human annotations needed for learning the instruction following task. 
Furthermore, it opens a new door for developing versatile and extremely sample-efficient embodied agents by harnessing the power of large language models and enhancing them with physical grounding.
Promising future directions include exploring other LLMs such as Codex~\cite{Chen2021EvaluatingLL}, better prompt design, and more advanced methods for grounding and dynamic re-planning.

\section*{Acknowledgement}

The authors would like to thank the colleagues from the OSU NLP group for their thoughtful comments. This research was supported by ARL W911NF2220144.

{\small
\bibliographystyle{ieee_fullname}
\bibliography{egbib}
}

\clearpage

\appendix

\section*{Appendices}

In this supplementary material, we present additional details and clarifications that are omitted in the main text due to space constraints.
\begin{itemize}
    \item \autoref{b:model_imp}: Additional Model Implementation Details
    \item \autoref{a:SL3}: Comparison with (SL)\textsuperscript{3} on ALFRED.
    \item \autoref{a:prompt_design}: Prompt design choices and prompt selection under true few-shot setting (cf. section 4.2 in the main paper).
    \item \autoref{a:additional_results}: Additional fine-grained analyses (cf. section 5 in the main paper).
\end{itemize}

\section{Additional Model Implementation Details}
\label{b:model_imp}

\subsection{HLSM} 
HLSM~\cite{blukis2021a} consists of three components: a semantic voxel map, a high-level planner, and a low-level planner. 
First, a 3D semantic voxel map is constructed by applying semantic segmentation and depth estimation to the visual inputs, which stores the agent's and the objects' real-time locations. 
Next, the high-level planner takes the language instructions, the semantic map encoding, and the previous subgoal history to predict the next subgoal. 
Lastly, the low-level planner is a mixture of deterministic algorithms and learned components (\eg, learning a yaw and pitch angle to face the object). 
HLSM first processes the sensory image input to create/update a map, which is used as an input to the high-level planner along with the language instructions to predict the next subgoal. 
Finally, the low-level planner maps the subgoal into a sequence of primitive actions.

To adapt HLSM to the few-shot setting, we need to re-train the components of the model that need paired trajectory-instruction data for training. 
For HLSM, paired data was only used for training the high-level controller. 
Therefore, we re-train the high-level controller with the same \num{100} training examples we use for \OurMethod. Specifically, we use the same set of hyperparameters as HLSM.
While the original HLSM focuses on the goal instruction only setting, we found that the step-by-step instructions are essential for the few-shot setting, so we concatenate goal instruction with step-by-step instructions for re-training HLSM's high-level planner. 
We leave the other components intact, which are downloaded from the official codebase.\footnote{\url{https://github.com/valtsblukis/hlsm}}

\subsection{FILM} FILM~\cite{min2022film} consists of four components: a semantic map, a semantic search policy, a template-based high-level planner, and a low-level planner. 
At the beginning of each task, five separate BERT-based classifiers~\cite{DBLP:conf/naacl/DevlinCLT19} are used to predict five parameters (task type, target objects, receptacles, parent objects, and whether slicing is needed), each of which takes the goal and optionally the step-by-step instructions as input to predict the respective parameter. 
FILM then generates the high-level plan by choosing a pre-defined template based on the predicted task type and filling the other parameters into the template. 
In addition, the semantic map is updated at each time step with the sensory image inputs.
At every \num{25} steps, the semantic search policy predicts the coordinates of the target object on the semantic map, which are then used by a deterministic low-level planner to decide on a low-level plan to navigate from the current location to the target object's location. 

Only the BERT-based classifiers need the language-related data for training.
Therefore, to adapt FILM to the few-shot setting, the five BERT-based classifiers are re-trained with the same \num{100} training examples used by the \OurMethod. 
Similar to HLSM, we concatenate the goal and the step-by-step instructions as input to the BERT-based classifiers. 
We use default hyperparameters for BERT models that are found in the paper.  
We use the predictions from these models to generate the high-level plans with the same pre-defined templates in FILM. 
We leave other components intact, which are downloaded from the official codebase.\footnote{\url{https://github.com/soyeonm/FILM}}

\begin{table*}[t]
    \centering
    \tabcolsep 5pt
    \renewcommand\arraystretch{1}
    \small
    \captionsetup{width=.85\textwidth} 
    \begin{tabular}{lcccccc}
        \toprule
       \textbf{Options} & \multicolumn{1}{c}{\textbf{\begin{tabular}[c]{@{}c@{}}Task\\ Introduction\end{tabular}}} & \multicolumn{1}{c}{\textbf{\begin{tabular}[c]{@{}c@{}}Goal\\ Instruction\end{tabular}}} & \multicolumn{1}{c}{\textbf{\begin{tabular}[c]{@{}c@{}}Step-by-step\\ Instructions\end{tabular}}} & \textbf{Plan List} & \textbf{Object List} & \multicolumn{1}{c}{\textbf{\begin{tabular}[c]{@{}c@{}}Retrieval\\ Message\end{tabular}}} \\
        \toprule
         \textbf{Default} & \begin{tabular}[c]{@{}l@{}}Create a high-level \\ plan for completing \\ a household task \\ using the allowed \\ actions and \\ visible objects.\\ \\ Allowed actions are \\ {[action list]} \end{tabular} & \begin{tabular}[c]{@{}l@{}}Task description: \\  {[goal instruction]}  \end{tabular}  & \begin{tabular}[c]{@{}l@{}}Step-by-step \\ instructions: \\ {[instructions]} \end{tabular}   & \begin{tabular}[c]{@{}l@{}}(Completed, Next) plan: \\ {[subgoals]} \end{tabular}   & \begin{tabular}[c]{@{}l@{}}Visible objects \\  are {[objects]} \end{tabular} & Next plan: \\
         \midrule
        \multirow{3}{*}{\textbf{Punctuation}} & ("PickupObject") & & & ("PickupObject", "Apple") & & \\
       & (PickupObject) & & & (PickupObject, Apple) & & \\ 
       & \textbf{PickupObject} & & & \textbf{PickupObject, Apple} & & \\
        \midrule 
        \multirow{3}{*}{\textbf{Naturalization}} & \textbf{PickupObject} & & & \textbf{PickupObject} & & \\
       & Pickup & & & Pickup & & \\
       & Pick up & & & Pick up & & \\
        \midrule
        \multirow{3}{*}{\textbf{Delimiter}} & & & \textbf{Pick up, go to} & \textbf{Pickup, Navigate} & \textbf{Apple, orange} &  \\
        & & & Pick up. Go to. & Pickup. Navigate & Apple. orange &  \\
        &  & & \multicolumn{1}{c}{\begin{tabular}[c]{@{}c@{}}Pick up \textbackslash n Go to \end{tabular}} & \multicolumn{1}{c}{\begin{tabular}[c]{@{}c@{}}Pickup \textbackslash n Navigate\end{tabular}}&
        \multicolumn{1}{c}{\begin{tabular}[c]{@{}c@{}}Apple \textbackslash n Orange\end{tabular}} & \\
    \bottomrule
    \end{tabular}
    \caption{For each element in our prompt design, we list the default phrasing. For the representation of actions, objects, and lists, we additionally experiment with different choices of punctuation, naturalization, and the delimiter between elements in a list.
    We select the optimal prompt design using LOOCV on the \num{100} training examples. The chosen options are highlighted in bold.}
    \label{tab:prompt_design}
\end{table*}

\begin{table}[t]
  \centering
  \small
  \renewcommand\arraystretch{1}
  \begin{tabular}{lcc}
    \toprule
     \multirow{2}{*}{\textbf{Task Type}} & \multicolumn{2}{c}{\textbf{HLP Accuracy}} \\
    \cmidrule(lr){2-2} \cmidrule(lr){3-3} &
    \textbf{Valid Unseen} & \textbf{Valid Seen}\\ 
    \midrule
    Pick \& Place & 51 & 46 \\
    Stack \& Place & 38 & 25 \\
    Place Two & 39 & 45 \\
    Examine & 44.4 & 49 \\
    Heat \& Place & 36 & 48 \\
    Cool \& Place & 43 & 46 \\
    Clean \& Place & 48.8 & 32  \\
    \bottomrule
  \end{tabular}
\caption{Static \OurMethod's high-level planning accuracy breakdown by task type.}
  \label{tab:hlp_error_task_type}
\end{table}

\subsection{SayCan}
SayCan~\cite{saycan2022arxiv} consists of 3 components: an LLM ranker, set of skills, and a value function. 
We use the LLM ranker adapted from SayCan's codebase~\footnote{\url{https://github.com/google-research/google-research/tree/master/saycan}} with the same settings (\eg temperature and log probability) and use GPT-3 (\fl{text-davinci-003}) as the choice of LLM.
First, SayCan generates a list of skills and their affordance score in the current environment using a pre-trained value function.
Then, it prompts the LLM with natural language description of each skill and generates a probability that represents how relevant it is to the task success.
Finally, SayCan combines the skill's LLM probability and the affordance score to choose which skill to execute.

To adapt SayCan to ALFRED, we need to define a \textit{skill} in the ALFRED environment.
From SayCan, a \textit{skill} is defined as \textit{“atomic” behaviors that are capable of low-level visuomotor control}.
Each skill can perform a short task, such as picking up a particular object.
This is identical to our definition of high-level plan in \S3, therefore we treat each skill as analogous to the (\fl{high-level action}, \fl{object}) pair.
This formulation allows us to use the same low-level controller we used for \OurMethod.
Furthermore, the value function is an another important concept for the  SayCan.
The value function predicts how likely an individual skill is to be executable in the current environment.
However, due to the resource constraint we were not able to generate the data and train a policy for the value function.
On the other hand, we decided to give SayCan an unfair advantage: we use the ground truth object information to construct an oracle value function.
Additionally, instead of iterating through a list of all possible (\fl{high-level action}, \fl{object}), we shrink the size of the skill to contain only the object type available in the current environment.
As we described in \S5.3, this gives SayCan an \textit{unfair competitive advantage} by giving it the oracle knowledge of all objects and affordances in the current environment \textit{a priori} to compiling the list of skills. 
Even though SayCan can shrink the skill space with the extra knowledge, SayCan's ranking nature calls LLM significantly more times than a generative model like \OurMethod. 
In fact, LLM-Planner calls GPT-3 avg.\ 7 times per task and SayCan calls it 22 times even with the oracle knowledge of the current environment to shrink the skill list.

\section{Comparison with (SL)\textsuperscript{3} on ALFRED}
\label{a:SL3}

(SL)\textsuperscript{3}~\cite{sharma-etal-2022-skill} is a recent hierarchical planning model that is also evaluated on the ALFRED benchmark.
It randomly samples \num{10}\% of ALFRED's training data for training. 
The high-level planner is based on a pre-trained T5-small~\cite{t5} model, which is fine-tuned to generate high-level plans from the goal instruction. 
The low-level planner is another fine-tuned T5-small model, which is tasked of generating a low-level plan for each subgoal in the high-level plan. 
Both goal and step-by-step instructions are needed for training, but only goal instructions are needed at inference time.

We could not compare (SL)\textsuperscript{3} under the same few-shot setting as \OurMethod because its code was not publicly available at the time of submission. 
However, we would like to highlight that our method achieves comparable performance on the validation set despite using only less than $1/20$ of training data than (SL)\textsuperscript{3} (\num{0.5}\% vs.\ \num{10}\% of ALFRED's training data).

\section{Prompt Design Choices}
\label{a:prompt_design}
In-context learning with GPT-3 could be sensitive to the prompt design. In \autoref{tab:prompt_design}, we show different prompt design choices we have experimented for \OurMethod. 
We structure our prompt into six consecutive parts: task introduction, goal instruction, step-by-step instruction, plan list, object list, and retrieval message. For each part, we have a default phrase and a list of additional options to try on top of the default phrasing signified as {[]}. All the options listed only modify the phrase that goes in {[]}.
First, we try adding punctuation marks around actions and object. 
Next, we naturalize each action name as a plain English text. Lastly, we experiment with finding the optimal delimiter between action list and step-by-step instruction list. We compared comma, period, and newline inserted between the sentences. The best prompt was chosen from the LOOCV accuracy for high-level plans and is bolded.

\section{Additional Fine-Grained Analyses}
\label{a:additional_results}

\subsection{HLP Accuracy by Task Type}

We show \OurMethod's high-level planning (HLP) accuracy breakdown by task type in \autoref{tab:hlp_error_task_type}. 
Because it is difficulty to determine a single value for the HLP accuracy for dynamic \OurMethod, here we focus on the static version, but the HLP accuracy of the dynamic version generally correlates well with that of the static version.
From the results, we observe that the results do not depend much on the difficulty of the task. For example, the task \nl{Stack \& Place} is often considered as the most difficult task based on the success rate of state-of-the-art models, but \OurMethod's HLP accuracy is similar to those of easier tasks such as \nl{Place two}. We find that \OurMethod is not overly sensitive to the complexity of tasks. This suggests that it could generalize well to different types of tasks with only a few in-context examples.

\begin{table}
  \centering
  \small
  \resizebox{1\linewidth}{!}{
  \renewcommand\arraystretch{1}
  \begin{tabular}{lcccccc}
    \toprule
      \textbf{Training Size} & \textbf{50} & \textbf{100} & \textbf{500} & \textbf{1k} & \textbf{10k} & \textbf{Full (21k)} \\ 
    \midrule 
    \OurMethod &  10.06  & 15.36   & 16.59  & 16.46  & 16.83  & 17.80  \\
    HLSM & 0.00  & 0.00 & 0.37  & 1.59 & 9.51  & 18.28  \\
    \bottomrule
  \end{tabular}
  }
\caption{Scaling experiment of \OurMethod and HLSM on valid unseen. Metric used is the task success rate.}
  \label{tab:hlp_scaling}
\end{table}

\begin{figure}[t]
    \centering
    \includegraphics[width=1\linewidth]{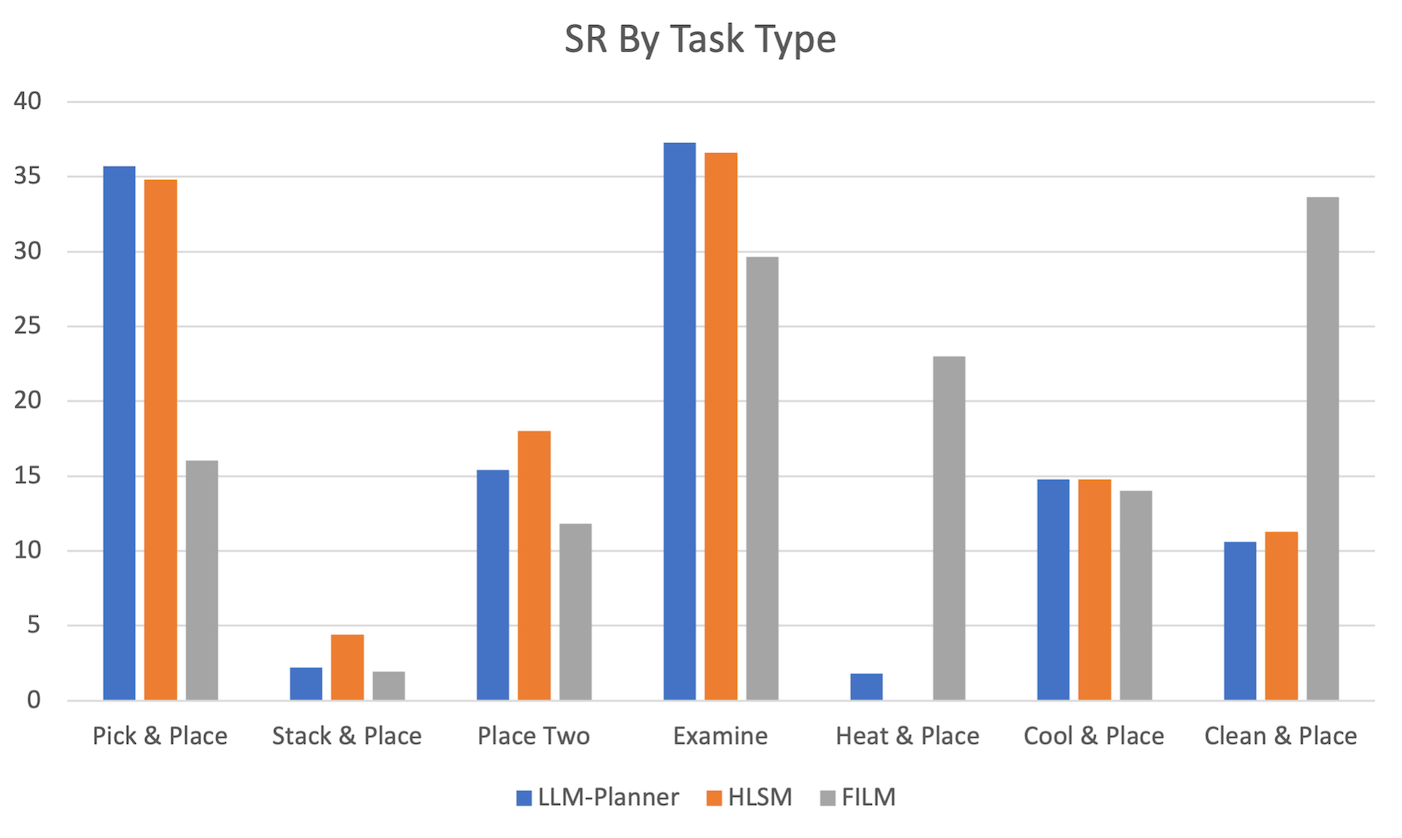}
    \caption{Success rate by task type on ALFRED valid unseen split.}
    \label{fig:sr_task_type}
\end{figure}

\subsection{End-to-End Performance by Task Type}

We show the end-to-end performance breakdown by task type of dynamic \OurMethod + HLSM in \autoref{fig:sr_task_type}. 
As a reference, we also compare with HLSM and FILM trained with the full training set of ALFRED.
Keep in mind that this is not apples-to-apples comparison because \OurMethod is under the few-shot setting.
Despite that, we can see that \OurMethod + HLSM achieves comparable performance with HLSM, and the distribution of the two are similar.
This is likely due to the shared low-level planner and object detector, which introduce a similar error profile. 
This again shows that our few-shot high-level planner is as good as HLSM's high-level planner that is trained with the full training set.
On the other hand, it also shows that there is still a large room to improve by using better low-level planners and object detectors.
For example, even though our HLP accuracy for \nl{Heat \& Place} is \num{36}\% as shown in \autoref{tab:hlp_error_task_type}, we could only get \num{1.8}\% success rate due to the object detector from HLSM often failing to detect the \nl{microwave}. 
If we use FILM's low-level planner and object detector, we may be able to achieve much better performance on this task.

\subsection{Scaling Comparison with HLSM}
We show \OurMethod's scaling experiments in comparison with the HLSM~\cite{blukis2021a} in ~\autoref{tab:hlp_scaling}.
We can see that \OurMethod significantly outperforms HLSM on almost all data size except for the full data setting.
This result shows that \OurMethod is more data-efficient across the board compared to the existing methods.
Even with the full data setting, \OurMethod only falls behind $0.48$ SR compared to the HLSM.
Our work can dramatically reduce the amount of human annotations needed for learning the task while maintaining a similar performance.

\end{document}